\documentclass[twocolumn]{svjour3}         
\smartqed  
\usepackage{graphicx}

\usepackage{titlesec} 
\titleformat{\paragraph}[runin]
{\bfseries\scshape}{\theparagraph}{1em}{}
\titlespacing{\paragraph}{0em}{1ex}{.5em} 
\usepackage{todonotes}
\usepackage{placeins} 
\usepackage{booktabs}
\usepackage{tabularx}
\usepackage{multicol}
\usepackage{bm}
\usepackage{hyperref}
\usepackage{url}
\usepackage{balance}
\usepackage{enumitem}
\usepackage{color}
\usepackage{bbding}
\usepackage{nccmath}
\usepackage{amsmath,amssymb}
\usepackage{cite}
\usepackage{makecell}
\usepackage{subfig}
\usepackage{epstopdf}
\usepackage{array}
\usepackage{multirow}
\usepackage{amsfonts}
\usepackage{float}
\usepackage{algorithm}
\usepackage{algpseudocode}
\usepackage[switch]{lineno}

\begin{document}

\title{Contrastive Decoupled Representation Learning and Regularization for Speech-Preserving Facial Expression Manipulation}


\author{Tianshui Chen \and
        Jianman Lin\and
        Zhijing Yang\thanks{Zhijing Yang is the corresponding author. Tianshui Chen and Jianman Lin contribute equally to this work and share co-first authorship. } \and
        Chumei Qing \and
        Yukai Shi \and
        Liang Lin
}


\institute{Tianshui Chen \at
             Guangdong University of Technology, Guangzhou, China \\
              \email{tianshuichen@gmail.com}
           \and
           Jianman Lin \at
           South China University of Technology, Guangzhou, China \\
              \email{linjianmancjx@gmail.com}
           \and
          Zhijing Yang \at
           Guangdong University of Technology, Guangzhou, China \\
              \email{yzhj@gdut.edu.cn}
            \and
          Chumei Qing \at
          South China University of Technology, Guangzhou, China \\
              \email{qchm@scut.edu.cn}
           \and
          Yukai Shi \at
           Guangdong University of Technology, Guangzhou, China \\
              \email{ykshi@gdut.edu.cn}
           \and
          Liang Lin \at
           Sun Yat-sen University, Guangzhou, China \\
              \email{linlng@mail.sysu.edu.cn}
}

\date{Received: date / Accepted: date}

\maketitle


\def\eg{\emph{e.g.}}
\def\etal{\emph{et al}}

\newcommand{\EQREF}{Eq.~\eqref}
\newcommand{\EQSREF}{Eqs.~\eqref}
\newcommand{\FIGREF}{Fig.~\ref}
\def\proposed{VB} 
\def\fixcolor{black}
\def\hstate{\bm {\tilde s}}
\def\rstate{\bm s}
\def\jstate{\bm s^{jn}}
\def\jpolicy{\overrightarrow{\pi}}
\def\vpref{v_{\text {pref}}}
\def\vector#1{\mbox{\boldmath $#1$}}
\def\sup#1{^{(\rm #1)}}
\def\sub#1{_{\rm #1}}
\def\supi#1{^{(#1)}}
\def\vct#1{\mbox{\boldmath $#1$}}
\def\eg{{\it e.g.}}
\def\cf{{\it c.f.}}
\def\ie{{\it i.e.}}
\def\etal{{\it et al. }}
\def\etc{{\it etc}}
\newcommand{\argmax}{\mathop{\rm argmax}\limits}
\newcommand{\argmin}{\mathop{\rm argmin}\limits}

\def\Rerr{\Delta \bm r}
\def\Terr{\Delta \bm t}
\def\Xerr{\Delta \bm x}
\def\XerrRel{\Delta \bm {\tilde x}}
\def\Xgt{\dot{\bm x}}
\def\Rgt{\dot{R}}
\def\Tgt{\dot{\bm t}}
\def\arraystretchlen{1.0}

\def\cam{c}
\def\image{\mathcal I}
\def\traj{\mathcal X}
\def\btraj{\mathcal {\bm X}}
\def\keypoints{\mathcal P}
\def\states{\mathcal S}
\def\bstates{\mathcal {\bm S}}
\def\state{\bm s}
\def\ped{\bm x}
\def\pedi{\bm p} 
\def\obs{\bm z}

\def\Fi{\bm F_r}
\def\Fp{\bm F_p}
\def\vpref{\bm w}
\def\ENERGY{{\mathcal E}}

\def\DIFF#1{\textcolor{black}{#1}}
\def\DIFFCR#1{\textcolor{black}{#1}}

\begin{abstract}
Speech-preserving facial expression manipulation (SPFEM) aims to modify a talking head to display a specific reference emotion while preserving the mouth animation of source spoken contents. Thus, emotion and content information existing in reference and source inputs can provide direct and accurate supervision signals for SPFEM models. However, the intrinsic intertwining of these elements during the talking process poses challenges to their effectiveness as supervisory signals. 
In this work, we propose to learn content and emotion priors as guidance augmented with contrastive learning to learn decoupled content and emotion representation via an innovative Contrastive Decoupled Representation Learning (CDRL) algorithm. Specifically, a Contrastive Content Representation Learning (CCRL) module is designed to learn audio feature, which primarily contains content information, as content priors to guide learning content representation from the source input. Meanwhile, a Contrastive Emotion Representation Learning (CERL) module is proposed to make use of a pre-trained visual-language model to learn emotion prior, which is then used to guide learning emotion representation from the reference input. We further introduce emotion-aware and emotion-augmented contrastive learning to train CCRL and CERL modules, respectively, ensuring learning emotion-independent content representation and content-independent emotion representation. 
During SPFEM model training, the decoupled content and emotion representations are used to supervise the generation process, ensuring more accurate emotion manipulation together with audio-lip synchronization. Extensive experiments and evaluations on various benchmarks show the effectiveness of the proposed algorithm.

\end{abstract}

\section{Introduction} 
\label{intro}
Speech-preserving facial expression manipulation (SPFEM) aims at manipulating facial emotions while maintaining mouth movements in static images or dynamic videos. It can significantly enhance human expressiveness and thus benefit various applications, ranging from virtual avatars to film and television production. For instance, current film-making often involves extensive efforts and multiple re-shoots to accurately capture an actor's intended emotions. In contrast, modifying facial emotions becomes simpler with a well-developed SPFEM system, offering similar outcomes in post-production. 

Current SPFEM works either adapt face reenactment algorithms \cite{tripathy2020icface,doukas2021head2head++} or proposes to modify latent representation \cite{papantoniou2022neural} to address the SPFEM task. The former works \cite{tripathy2020icface,doukas2021head2head++} manipulates facial expressions through the exchange of latent codes \cite{karras2019style} or facial action units \cite{friesen1978facial}, and employs the reference images as surrogate labels to construct frame-by-frame construction supervision. However, the reference images are not perfect targets as it exhibit mouth animation of original content, leading to generating sub-optimal results. The latter works \cite{papantoniou2022neural} propose to replace 3DMM parameters \cite{blanz2023morphable,filntisis2023spectre,tewari2020stylerig,ding2023diffusionrig} (i.e., exp parameters) with those from reference images to modify the expression. Despite achieving better performance, the 3DMM parameters in the mouth area are inherently intricately interwoven with other facial parameters, compromising preserving the mouth animations of spoken content. These works can not achieve expressive motion and accurate lip-sync simultaneously since spoken content and expression are intrinsically intertwined during the talking process. Thus, it is crucial to decouple content and emotional information from source and target images, which can be served as more direct and accurate supervision signals. 

In this work, we introduce a novel Contrastive Decoupled Representation Learning (CDRL) algorithm, which learns decoupled content and emotional representation as additional supervision signals for SPFEM model training. \textcolor{black}{Specifically, we first design a Contrastive Content Representation Learning (CCRL) module to exploit audio clip of the source input, which mainly refers to information of spoken contents, as content prior to guide learning content representation via a cross-attention mechanism. To ensure excluding emotional information, we further introduce an emotion-aware contrastive loss to train the CCRL module, which maximizes the similarities between content representations of inputs expressing identical audio content with different emotions while minimizing the similarities between those of inputs expressing different audio content with the same emotion. Meanwhile, we propose a Contrastive Emotion Representation Learning (CERL) module that exploits a pre-trained visual-language model \cite{radford2021learning} with prompt tuning to learn emotion priors. These priors are then used to guide learning emotion representation via a simple correlation operation.} \textcolor{black}{Recognizing that different emotions often share overlapping characteristics, we design an emotion-augmented contrastive loss that selectively employs samples with high emotional clarity to train the CERL module, ensuring the capture of accurate emotional information.} During SPFEM model training, we pose consistency constraints between content representations of generated image and source input and that between emotion representation of generated image and reference input. 

The contributions can be summarized into four folds. Firstly, we introduce a CDRL algorithm that learns decoupled content and emotion representation as a more direct and accurate supervision signal for SPFEM model training. To our knowledge, this is the first attempt to explicitly decouple content and emotional information from talking head videos to facilitate the SPFEM task. Second, we design a Contrastive Content Representation Learning (CCRL) module that combines a cross-attention mechanism with emotion-aware contrastive loss to learn emotion-independent content representation. Third, we design a Contrastive Emotion Representation Learning (CERL) module that exploits prompt tuning of large-scale visual-language models equipped with emotion-augmented contrastive learning to learn content-independent emotion representation. Finally, we conduct extensive experiments on various benchmarks, demonstrating that the proposed algorithm can better preserve the audio-lip synchronization and manipulate emotional states.

\section{Related Works}
\subsection{Facial Expression Manipulation}
Facial expression manipulation involves altering facial expressions in images or videos using various image-to-image translation methods. Several methods have been developed for this purpose, including those by \cite{dalva2023image, isola2017image, zhu2017unpaired, choi2018stargan, ververas2020slidergan}. Additionally, there are specific methods for facial expression editing, such as \cite{yang2024ccr, liu2023gan, tripathy2020icface, ding2018exprgan, geng20193d, tewari2020stylerig, d2021ganmut, xu2023progressive, kollias2020deep, geng2020towards}. For instance, ExprGAN \cite{ding2018exprgan} is a method based on conditional GANs, enabling transformation of faces into specified expressions with continuous intensities. GANmut \cite{d2021ganmut} introduces a GAN-based framework that learns an expressive and interpretable conditional space of emotions. GANimation \cite{pumarola2020ganimation} uses adversarial learning conditioned on action unit (AU) annotations \cite{friesen1978facial} to describe facial movements in a continuous manifold, allowing control over the activation magnitude of each AU and the combination of multiple AUs. Head2Head++ \cite{doukas2021head2head++} employs a sequential generator and a customized dynamics discriminator to achieve temporally consistent video manipulation. While these methods achieve impressive results in transforming facial expressions, they struggle to simultaneously transform emotion-related expressions while retaining lip synchronization. Specifically, translating the expression of the speaker in each frame often changes the mouth shape due to biases in the training data distribution.

Recently, StyleGAN-based expression manipulation has gained attention due to the semantically disentangled latent spaces of StyleGAN and the high quality of the generated results \cite{karras2019style, karras2020analyzing}. This process begins by projecting the input image into StyleGAN's latent space \cite{xia2022gan}. This projection can be achieved either through optimization-based methods \cite{roich2022pivotal, karras2020analyzing, lipton2017precise, abdal2019image2stylegan, abdal2020image2stylegan++} or encoder-based methods \cite{richardson2021encoding, tov2021designing, alaluf2021restyle, zhu2024domain, hu2022style, liu2023fine, li2023reganie, pehlivan2023styleres, cao2024decreases, yang2023out, yildirim2023diverse, wang2022high, xu2024self}. After the input image is projected into the latent space, the corresponding latent code is adjusted towards the location of the target emotion. Finally, StyleGAN generates the edited image from this modified latent code. A representative method, PTI \cite{roich2022pivotal}, first identifies a pivot latent code to approximate the input image. It then fine-tunes the generator's weights to enhance the reproduction of the target image and facilitate image manipulation. To achieve semantically consistent continuous editing together with temporal consistency, STIT \cite{tzaban2022stitch} recovers original temporal correlations by faithfully inverting each frame. It fine-tunes a unique generator for each input video, enabling the generator to capture all reconstruction details. Building on STIT, TCSVE \cite{xu2022temporally} introduces a temporal consistency loss for edited videos, enhancing the temporal coherence of the results. However, both STIT and TCSVE are video-specific, requiring retraining for each new video. This leads to high training costs and limited generalization ability. \textcolor{black}{In contrast, RIGID \cite{xu2023rigid} addresses these limitations by learning the inherent coherence between input frames in an end-to-end manner. This approach makes it agnostic to specific emotions and applicable to arbitrary editing of the same video without the need for retraining.} Although the StyleGAN-based expression manipulation method can achieve speech preserving and temporal consistency in facial expression editing, it faces two major challenges: finding distinguishable and decoupled editing directions for different emotions and correctly embedding each frame of the video into the StyleGAN latent space to achieve high-fidelity editing. Both processes are very time-consuming, limiting the applicability of StyleGAN-based emotion manipulation methods in real-world scenarios.

\begin{figure*}[htp]
  \centering
  \includegraphics[width=1.0\textwidth]{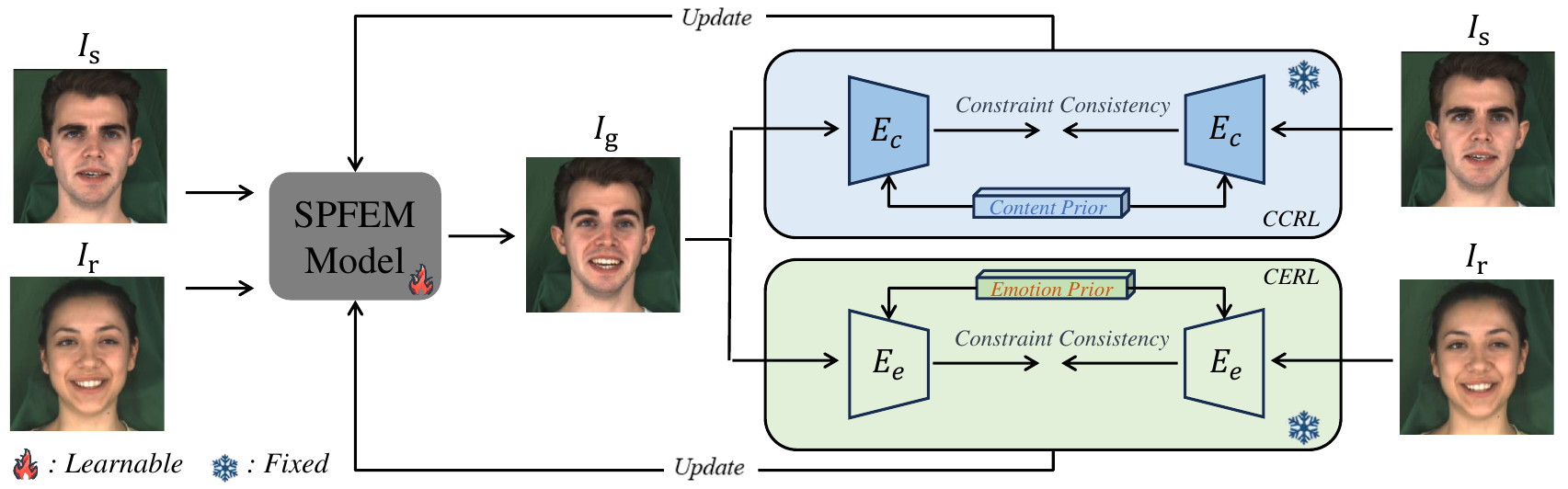} 
  \vspace{-20pt}
  \caption{An overall pipeline of incorporating the proposed CDRL algorithm to supervise learning SPFEM models. It consists of the CCRL and CERL modules. \textcolor{black}{CCRL utilizes the audio corresponding to the source input ($I_s$) as content prior to decoupling content representation from both the source input ($I_s$) and the generated output ($I_g$), ensuring aligned content generation. CERL employs the learned emotion prior for decoupling emotions from the reference input ($I_r$) and the generated output ($I_g$), facilitating consistent emotion generation.}}
  \label{fig: framework}
\end{figure*}

\subsection{Speech-Preserving Facial Expression Manipulation}
The SPFEM Model aims to alter the given source video to display the desired emotion while preserving the facial animation corresponding to the voice content. Unlike StyleGAN-based facial expression editing, the SPFEM model is neither video-specific nor emotion-specific. Once trained, it can be applied to any video and any emotion modification for the same speaker. ICface \cite{tripathy2020icface} controls the pose and expression with interpretable control signals such as head pose angles and action units. The Wav2Lip-Emotion method \cite{2021Invertible} extends the lip synchronization architecture \cite{prajwal2020lip} by modifying facial emotion using L1 reconstruction and pre-trained emotion objectives. However, both methods struggle to preserve facial identity in test images, and the visual quality of the generated images is very low.

3D Morphable Models (3DMM) \cite{blanz2023morphable, filntisis2023spectre, feng2021learning, tewari2020stylerig, ding2023diffusionrig, fu2024mimic, ververas2020slidergan} explicitly model facial movements. Additionally, \cite{sun2023continuously} demonstrate that 3DMM can capture large-scale deformations such as opening the mouth wide in anger or raising eyebrows in joy, influencing the perception of whether an expression is positive or negative. These characteristics make 3DMMs particularly well-suited for use in the SPFEM task. DSM \cite{solanki2023dsm} enables semantic video manipulation using neural rendering and 3DMM, providing intuitive control of facial expressions and introducing an AI tool that maps semantic labels to the Valence-Arousal space, translating them into photorealistic 3D facial expressions. NED \cite{solanki2023dsm} proposed a framework based on a parametric 3D face representation that disentangles facial identity from head pose and expressions. It uses deep domain translation to consistently alter facial expressions and a neural face renderer for photorealistic manipulation. Recognizing that 3DMM cannot capture color changes and some fine facial details such as wrinkles, \cite{sun2023continuously} present a new approach for this task as a special case of motion information editing. They use a 3DMM to capture major facial movements and an associated texture map modeled by a StyleGAN to capture appearance details, which is more effective in achieving photorealistic and detailed facial expression manipulation.

Despite these methods making significant progress, a key limitation is the lack of paired supervision, which has led to suboptimal outcomes in both emotion manipulation and the preservation of speech content. In contrast, our work introduces a novel Contrastive Decoupled Representation Learning (CDRL) algorithm. This approach focuses on separately learning content and emotional representations, subsequently integrating these independently refined elements as supervisory signals during the training process of the SPFEM model, offering a more effective solution.

\section{Method}
In this section, we introduce the CDRL algorithm, which consists of CCRL and CERL modules. CCRL exploits audio as content prior to guiding learning emotion independent content representation from the source images while CERL first introduces a visual-language model to learn emotion priors and uses these priors to guide learning content-independent emotion representation. To ensure capture of the content and emotion representations, we introduce emotion-aware and emotions-augmented contrastive learning to train these two modules, respectively. During SPFEM model training, content and emotion representations are used as more direct and accurate supervision signals. An overall illustration of incorporating the CDRL algorithm into the SPFEM models is presented in Fig. \ref{fig: framework}

\begin{figure}[!t]
    \centering
    \includegraphics[width=0.48\textwidth]{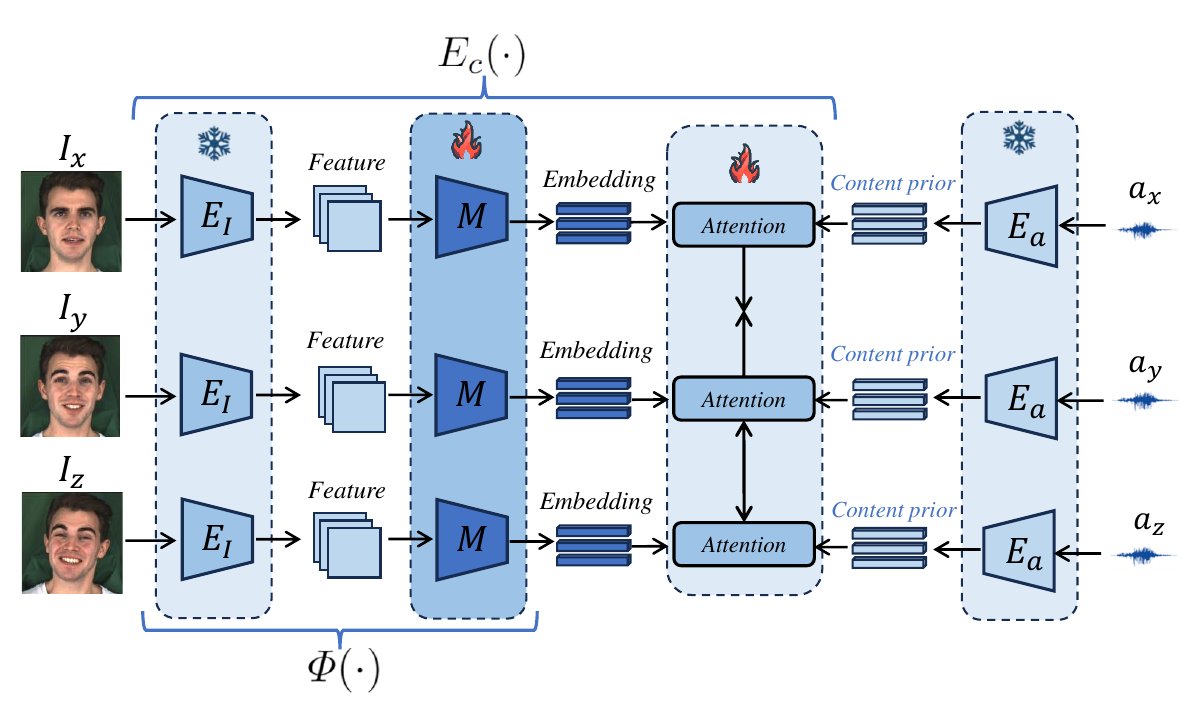}
    \vspace{-20pt}
    \caption{An illustration of CCRL module. It utilizes the audio clip to guide learning content representation through a cross-attention mechanism equipped with an emotion-aware contrastive loss. \textcolor{black}{In this context, The image encoder $\Phi(\cdot)$ combines the pretrained ArcFace $E_I(\cdot)$ \cite{deng2019arcface} and the mapping operation $M(\cdot)$, while $E_c(\cdot)$ consists of $\Phi(\cdot)$ and a cross-attention mechanism.}}
    \label{fig: CCRL}    
\end{figure}

\subsection{Contrastive Content Representation Learning}
CCRL first employs a cross-attention mechanism \cite{vaswani2017attention}, which exploits audio information to guide focusing on content-related features. Then, it uses emotion-aware contrastive loss to further exclude the emotional information. 

Formally, given three image frames $I_x$, $I_y$, $I_z$ and their corresponding audio clips $a_x$, $a_y$, $a_z$, in which $I_x$, $I_y$ enjoy identical speech content and have different emotions while $I_y$, $I_z$ expressing different speech contents but exhibiting the same emotion, we utilize an image encoder to extract image features and an audio encoder to extract audio features. Since the audio features mainly contain information related to spoken content, it is expected to use it to guide focusing on content-related areas and thus extract content representation. Here, we introduce the cross-attention mechanism that treats audio features as query and image features as key and value to achieve this end, formulated as

\begin{align}
\begin{split}
f_{xa} & = \text{Cross\_Att}(\Phi(I_x), E_a(a_{x})) \\ 
f_{ya} & = \text{Cross\_Att}(\Phi(I_y), E_a(a_{y})) \\ 
f_{za} & = \text{Cross\_Att}(\Phi(I_z), E_a(a_{z})) 
\end{split}
\end{align}

Where image encoder $\Phi(\cdot)$ is implemented by a pretrained ArcFace $E_I(\cdot)$ \cite{deng2019arcface} followed by a learnable mapping operation $M(\cdot)$. Audio encoder $E_a(\cdot)$ is implemented by the pretrained XLSR \cite{conneau2020unsupervised}. To ensure CCRL only focuses the content information, we further introduce an emotion-aware contrastive loss inspired by the recent progress in previous works \cite{chen2020simple,wu2022ccpl,chen2024heterogeneous}, Formally

\begin{equation}
\mathcal{L}_{ccrl} = \sum_{(x, y) \in P} \left( (1 - \varphi (f_{xa}, f_{ya})) \right) + \sum_{(y, z) \in N} \left( \varphi (f_{ya}, f_{za}) \right)
\end{equation}

Where $\varphi$ is employed to denote the cosine similarity function, $f_{xa}$ and $f_{ya}$ are mutually positive samples, whereas $f_{ya}$ and $f_{za}$ serve as negative samples. The image sets of positive and negative samples are denoted as $P$ and $N$, respectively. Considering our specific context where images $I_x$, $I_y$, and $I_z$ originate from the same speaker, image $I_z$ is characterized by differing content but identical emotional information compared to image $I_y$ while differing in both aspects from image $I_x$. By designating $f_{ya}$ and $f_{za}$ as negative samples, this approach achieves two key objectives: (1). It ensures that cross-attention mechanisms are not biased by the identity information; (2). Since $f_{ya}$ and $f_{za}$ share the same emotional information, their classification as negative samples further aids in the decoupling of content from emotional information. Once learned, the audio feature is considered as content prior to decoupling emotion-independent content representation for supervised content generation in SPFEM models as shown in Fig \ref{fig: framework}.

\subsection{Contrastive Emotion Representation Learning}
CERL initially utilizes a pre-trained visual-language model \cite{radford2021learning} coupled with prompt tuning, to learn emotion priors, guiding the focus towards emotion-related features. Subsequently, it employs an emotion-augmented contrastive loss to further emphasize the exclusion of other information.

\begin{figure}[!t]
    \centering
    \includegraphics[width=0.48\textwidth]{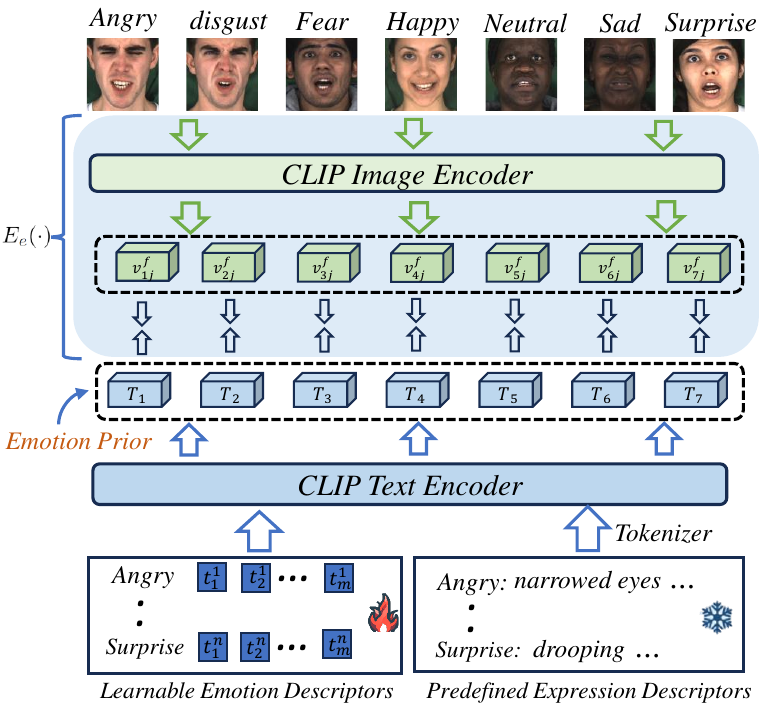}
    \caption{An illustration of CERL module. It uses a pre-trained visual-language model with prompt tuning to learn emotion priors and exploits the priors to guide learning emotion representation with a simple correlation operation supervised by an emotion-augmented contrastive loss. \textcolor{black}{$E_e(\cdot)$ includes image feature extraction and a dot product with the emotion prior.}}
    \label{fig: CERL}
\end{figure}

Inspired by recent advances in visual-language models like CLIP, there is a strong interest in utilizing these models to extract emotional representation from images. We conducted a zero-shot emotion classification experiment using seven distinct emotion labels to assess CLIP's capability. Remarkably, CLIP demonstrated proficiency with emotionally expressive images from the MEAD dataset, achieving classification scores exceeding 0.8. This highlights CLIP's potential in discerning and capturing emotional semantics within images. Building on this, we introduce the Contrastive Emotion Representation Learning (CERL) module, which learns emotion priors for each emotional state via prompt tuning together with emotion-augmented contrastive learning, as depicted in Fig \ref{fig: CERL}.

The concept of emotion is represented using eight placeholders ``[C]'', each associated with a learnable vector $t_{m}^{n}$, where $n$ $\in [1,7]$ and $m \in [1,8]$. This signifies seven distinct emotion categories, each with eight unique placeholder characters ``[C]''. Additionally, predefined generic emotion descriptions serve as auxiliary information \cite{zhao2023prompting}. These descriptions are combined with $t_{m}^{n}$ and processed through CLIP's text encoder to generate the emotion prior $T_n$, which is the primary focus of our learning process.

\textcolor{black}{Recognizing that different emotions often share overlapping characteristics, we selected a large number of the most expressive images for each emotion from the MEAD dataset to explore the subtle differences underlying each emotion.} Specifically, CLIP serves as a filtering tool for selecting emotionally expressive images for each emotion. It uses seven emotion classification labels, each with a corresponding threshold value, to selectively filter images based on their emotional expressiveness. This process results in the creation of seven sub-datasets denoted as $D_n$, each corresponding to a specific emotion. To extract image features from these sub-datasets, we integrate CLIP's image encoder into CERL. The extracted features of the $j$ th image in subset $D_n$ are represented as $v_{n,j}^{f}$. During training, we utilize an emotion-augmented contrastive learning strategy, treating matching pairs of $v_{n,j}^{f}$ and $T_n$ with the same emotion as positive samples, while pairs with different emotions are considered negative samples. This process distills emotion priors from images with the most significant emotional representations:

\begin{align}
\begin{split}
\mathcal{L}_{\text{cerl}} = -\sum_{i=1}^{n} \sum_{j=1}^{r} \log 
\frac{ \exp\left( \frac{T_i v_{i,j}^{f}}{\tau} \right) }
{ \exp\left( \frac{T_i v_{i,j}^{f}}{\tau} \right) 
+ \sum_{\substack{k=1 \\ k \ne i}}^{n} 
\exp\left( \frac{T_i v_{k,j}^{f}}{\tau} \right) }
\end{split}
\end{align}

Here, $r$ represents the number of images within the sub-dataset. Through the training of the CERL, we can derive seven distinct emotions prior $T_n$, which are distilled from a vast dataset comprising thousands of images, capturing universal emotions prior that are independent of ID information and content information. As illustrated in Fig \ref{fig: framework}, we use $T_n$ to assist in acquiring the content-independent emotion representation to supervise emotion generation in the SFPEM model.

\subsection{Content  and Emotion Regularization} 
\textcolor{black}{In the previous section, we detailed the training of the CCRL and CERL modules. We now turn our attention to their integration into the SPFEM model. During SPFEM model training, the pre-trained CCRL module is used to compute the content regularization loss between the source input and the generated output, while the pre-trained CERL module calculates the emotion regularization loss between the reference input and the generated output. These two regularization terms are then weighted and combined, providing an additional signal to guide the training process of the SPFEM model through backpropagation.}

Formally, the content source, emotion reference, and SPFEM model's output are denoted as $I_s$, $I_r$, and $I_g$, respectively. We define $E_c(\cdot)$ as the combination of the image encoder $\Phi(\cdot)$ and a cross-attention mechanism. Briefly, audio $a_s$ is extracted from $I_s$, with $E_a(\cdot)$ transforming it into a content prior. Then, $E_c(\cdot)$ processes $I_s$, $I_g$, and this content prior, initially extracting image features via $\Phi(\cdot)$ and subsequently merging them with the content prior to acquiring the decoupled emotion-independent content representation

\begin{align}
\begin{split}
f_{sa} &= E_c(I_s, E_a(a_s)) \\
f_{ga} &= E_c(I_g, E_a(a_s)) \\
\mathcal{L}_{\text{c}} &= 1 - \varphi(f_{sa}, f_{ga})
\end{split}
\end{align}

By maximizing the similarity between the decoupled emotion-independent content representations \( f_{sa} \) and \( f_{ga} \), we can regularize the content generation of \( I_g \) using the content source \( I_s \).

For the regularization of generating emotional information, we utilize learned $T_n$ as the emotion prior. To maintain the alignment of the visual and textual information in the embedding space of CLIP, we simply employ a dot product to produce the emotional representation between the image and emotion prior

\begin{align}
\begin{split}
\mathcal{L}_{\text{e}} = 1 - \varphi(E_e(I_r, T_i), E_e(I_g, T_i))
\end{split}
\end{align}

Where $T_i$ is the emotion prior to the corresponding emotion of $I_r$, $E_e(\cdot)$ encompasses image feature extraction and the dot product of image features with the emotion prior to capturing content-independent emotional representations. We employ the $\mathcal{L}_{\text{e}}$ to regularize the emotion generation process of the SPFEM model.

Current SPFEM algorithms can be categorized into two types. The first type, exemplified by NED \cite{papantoniou2022neural}, follows a two-stage generation process. In this approach, the first stage generates 3DMM parameters, and the second stage utilizes these parameters to render the final images. The second type, represented by ICface \cite{tripathy2020icface}, directly generates the rendered images. 

\begin{align}
\begin{split}
\mathcal{L}_{CDRL} = \mathcal{L}_{\text{c}} + \mathcal{L}_{\text{e}}
\end{split}
\end{align}

The visual information can pertain to either the intermediate 3DMM parameters or the final rendered images, and our algorithm can use $\mathcal{L}_{CDRL}$ to regularize both. \textcolor{black}{\emph{The integration of $\mathcal{L}_{CDRL}$ into the training pipeline of NED and ICface are detailed in the supplementary material.}}


\section{Experiments}
\subsection{Experiment Settings}
\noindent\textbf{Dataset. } MEAD \cite{wang2020mead} contains 60 speakers, and each speaker records 30 videos in each emotional state (i.e., neutral, happy, angry, surprised, fear, sad, and disgusted). Here, we select the videos of 36 speakers that have 7,560 videos to train the CCRL and CERL modules. To evaluate their performance, we integrated them into existing SPFEM models, ICface and NED. For the evaluation, we chose six distinct speakers (M003, M009, W029, M012, M030, and W015) who collectively contributed 1,260 videos. In line with prior research methodologies, we randomly allocated 90\% of these videos to the training set and reserved the remaining 10\% for the test set.  Additionally, we conducted tests on the RAVDESS dataset \cite{livingstone2018ryerson}, applying the CCRL and CCRL modules without any retraining. For this, we selected six speakers (actors 1-6), who collectively contributed a total of 168 videos. Consistent with our previous methodology, we randomly assigned 90\% of these videos to the training set and utilized the remaining 10\% as the test set.

\noindent\textbf{Evaluation Protocol. }
In our study, we employ the following evaluation metrics: 1) Frechet Arcface Distance (FAD) measures the realism, with a low FAD denoting better realism \cite{deng2019arcface}. 2) Cosine Similarity (CSIM) evaluates emotion similarity, with a higher CSIM indicating high similarity. 3) Lip Sync Error Distance (LSE-D) computes lip-audio synchronization, with lower LSE-D values reflecting superior lip-audio precision \cite{prajwal2020lip}. Our results are showcased across two settings: intra-ID, featuring the same speaker in both emotion reference and source video, and cross-ID, where the speakers differ.

\begin{table*}[!t]
\scriptsize
\centering
\resizebox{0.98\textwidth}{!}{
\begin{tabular}{c|c|ccc|ccc|ccc|ccc}
\toprule
\multirow{2}{*}{Settings} &  \multirow{2}{*}{Emotions} &  \multicolumn{3}{c|}{ICface}  & \multicolumn{3}{c|}{Ours (ICface)}  & \multicolumn{3}{c|}{NED}  & \multicolumn{3}{c}{Ours (NED)} \\
\cline{3-14}
& & FAD$\downarrow$ & LSE-D$\downarrow$ & CSIM$\uparrow$ & FAD$\downarrow$ & LSE-D$\downarrow$ & CSIM$\uparrow$ &  FAD$\downarrow$ & LSE-D$\downarrow$ & CSIM$\uparrow$ &  FAD$\downarrow$ & LSE-D$\downarrow$ & CSIM$\uparrow$ \\
\hline
\multirow{8}{*}{intra-ID} 
 & Neutral    & 7.114 & 9.760  & 0.779 & 7.251 & 9.281 & 0.784 & 0.906 & 9.264  & 0.883 & 0.722 & 9.255  & 0.916 \\
 & Angry      & 6.420 & 10.483 & 0.741 & 6.199 & 9.362 & 0.801 & 2.177 & 9.579  & 0.802 & 1.045 & 9.682  & 0.896 \\
 & Disgusted  & 7.383 & 10.433 & 0.805 & 6.265 & 9.266 & 0.815 & 3.838 & 9.128  & 0.772 & 1.115 & 9.213  & 0.927 \\
 & Fear       & 6.567 & 9.855  & 0.754 & 6.696 & 9.389 & 0.812 & 1.659 & 10.172 & 0.848 & 1.228 & 9.490  & 0.908 \\
 & Happy      & 6.213 & 10.180 & 0.775 & 6.198 & 9.379 & 0.827 & 1.939 & 9.137  & 0.839 & 0.987 & 9.427  & 0.923 \\
 & Sad        & 7.301 & 10.017 & 0.755 & 6.707 & 9.398 & 0.792 & 2.538 & 9.074  & 0.812 & 1.258 & 9.243  & 0.911 \\
 & Surprised  & 6.567 & 9.851  & 0.817 & 7.438 & 9.290 & 0.798 & 1.700 & 9.821  & 0.864 & 0.825 & 9.327  & 0.918 \\
\cline{2-14}                                            
 & Avg. & 6.795 & 10.083 & 0.775 & 6.679 & 9.338 & 0.804 & 2.108 & 9.454 & 0.831 & 1.026 & 9.372 & 0.914 \\
\hline
\hline
\multirow{8}{*}{Cross-ID} 
 & Neutral    & 10.560 & 11.226 & 0.705 & 9.976 & 10.423 & 0.681 & 2.022 & 9.812  & 0.841 & 1.995 & 9.393 & 0.849 \\
 & Angry      & 9.470  & 11.073 & 0.648 & 8.573 & 10.165 & 0.667 & 4.851 & 9.904  & 0.717 & 4.988 & 9.307 & 0.740 \\
 & Disgusted  & 9.230  & 11.184 & 0.637 & 8.558 & 10.633 & 0.785 & 5.094 & 10.121 & 0.791 & 4.687 & 9.292 & 0.814 \\
 & Fear       & 9.122  & 11.204 & 0.727 & 9.279 & 10.163 & 0.720 & 4.983 & 9.741  & 0.750 & 5.021 & 9.456 & 0.767 \\
 & Happy      & 8.493  & 11.322 & 0.717 & 8.837 & 10.055 & 0.795 & 3.919 & 9.936  & 0.842 & 3.264 & 9.297 & 0.870 \\
 & Sad        & 10.364 & 11.526 & 0.664 & 9.073 & 10.608 & 0.689 & 5.665 & 10.179 & 0.691 & 5.479 & 9.353 & 0.712 \\
 & Surprised  & 9.541  & 11.133 & 0.721 & 10.197 & 10.338 & 0.743 & 4.600 & 9.646  & 0.780 & 4.976 & 9.362 & 0.793 \\
\cline{2-14} 
 & Avg. & 9.540 & 11.238 & 0.688 & 9.213 & 10.341 & 0.726 & 4.448 & 9.906 & 0.773 & 4.344 & 9.351 & 0.792 \\
\hline
\end{tabular}}
\caption{Comparision results of FAD, CSIM, and LSE-D of NED, ICface with and without our CDRL on the intra-ID and cross-ID settings on the MEAD dataset.}
\label{table: MEAD}
\end{table*}

\begin{table*}[!t]
\scriptsize
\centering
\resizebox{0.98\textwidth}{!}{
\begin{tabular}{c|c|ccc|ccc|ccc|ccc}
\toprule
\multirow{2}{*}{Settings} &  \multirow{2}{*}{Emotions} &  \multicolumn{3}{c|}{ICface}  & \multicolumn{3}{c|}{Ours (ICface)}  & \multicolumn{3}{c|}{NED}  & \multicolumn{3}{c}{Ours (NED)} \\
\cline{3-14}
& & FAD$\downarrow$ & LSE-D$\downarrow$ & CSIM$\uparrow$ & FAD$\downarrow$ & LSE-D$\downarrow$ & CSIM$\uparrow$ &  FAD$\downarrow$ & LSE-D$\downarrow$ & CSIM$\uparrow$ &  FAD$\downarrow$ & LSE-D$\downarrow$ & CSIM$\uparrow$ \\
\hline
\multirow{8}{*}{intra-ID} 
& Neutral    & 9.816& 8.209& 0.749 & 7.589 & 7.041  & 0.765 & 2.041& 7.376& 0.847 & 2.761& 7.321& 0.859 \\
& Angry      & 7.047& 9.504& 0.703 & 5.866 & 10.122 & 0.709 & 3.288& 7.757& 0.805 & 3.721& 7.502& 0.789 \\
& Disgusted  & 8.689& 8.295& 0.775 & 6.497 & 9.199  & 0.795 & 4.144& 7.822& 0.786 & 3.189& 7.779& 0.839 \\
& Fear       & 8.413& 8.523& 0.722 & 6.780 & 9.478  & 0.745 & 2.635& 7.452& 0.842 & 2.489& 7.821& 0.836 \\
& Happy      & 8.413& 8.902& 0.797 & 7.007 & 8.130  & 0.781 & 3.714& 7.742& 0.793 & 3.031& 6.567& 0.829 \\
& Sad        & 8.086& 8.346& 0.766 & 6.827 & 7.377  & 0.796 & 2.595& 7.560& 0.855 & 2.266& 7.112& 0.849 \\
& Surprised  & 8.636& 7.578& 0.772 & 7.127 & 7.497  & 0.793 & 2.980& 7.226& 0.848 & 3.404& 7.312& 0.860 \\
\cline{2-14}                                                                                       
& Avg. & 8.443& 8.480 & 0.755 & 6.813 & 8.406 & 0.769 & 3.057& 7.562& 0.825  & 2.980 & 7.345 & 0.837 \\
\hline
\hline
\multirow{8}{*}{Cross-ID} 
& Neutral   & 10.478 & 10.736 & 0.677 & 9.198 & 8.542  & 0.669 & 3.558 & 7.856 & 0.820 & 3.162 & 7.551 & 0.809 \\
& Angry     & 8.704  & 12.415 & 0.646 & 7.744 & 12.429 & 0.645 & 5.546 & 8.085 & 0.766 & 4.852 & 8.492 & 0.742 \\
& Disgusted & 9.260  & 11.860 & 0.717 & 7.168 & 11.655 & 0.715 & 7.388 & 8.107 & 0.741 & 7.541 & 7.931 & 0.749 \\
& Fear      & 9.106  & 11.279 & 0.649 & 8.838 & 11.36  & 0.658 & 5.008 & 8.151 & 0.749 & 4.061 & 7.728 & 0.797 \\
& Happy     & 9.061  & 11.150 & 0.738 & 8.326 & 9.486  & 0.744 & 5.648 & 8.073 & 0.804 & 4.819 & 7.943 & 0.799 \\
& Sad       & 9.639  & 11.305 & 0.666 & 8.487 & 8.347  & 0.686 & 5.588 & 8.006 & 0.726 & 4.849 & 7.521 & 0.739 \\
& Surprised & 9.718  & 12.028 & 0.644 & 9.191 & 8.866  & 0.633 & 5.145 & 7.962 & 0.713 & 4.429 & 7.596 & 0.759 \\
\cline{2-14}                                                                      
& Avg. & 9.424 & 11.539 & 0.677 & 8.422 & 10.098 & 0.679 & 5.412 & 8.034 & 0.760  & 4.816 & 7.823 & 0.771  
\\
\hline
\end{tabular}}
\caption{Comparision results of FAD, CSIM, and LSE-D of NED, ICface with and without our CDRL on the intra-ID and cross-ID settings on the RAVDESS dataset.}
\label{table: RAVDESS}
\end{table*}

\subsection{Implementation Details}
\noindent\textbf{Paired Data Construction. }
We utilize the MEAD dataset as the foundation for training our CDRL algorithm. Despite the presence of videos within the MEAD that feature a speaker uttering the same sentence in diverse emotional states, acquiring pairs of image data where an image of a sentence spoken in one emotional state corresponds to another image of the same sentence spoken in a different emotional state remains challenging. To address this, we employ the Dynamic Time Warping (DTW \cite{berndt1994dtw}) algorithm to align the Mel spectra of the two videos, thereby obtaining one-to-one training data. This paired data is then utilized to train the CDRL algorithm.

\noindent\textbf{CCRL. }
During the training phase of CCRL, the network architecture maintains fixed parameters for both ArcFace \cite{deng2019arcface} and XLSR \cite{conneau2020unsupervised}, focusing the training efforts specifically on the cross-attention mechanism and the module $M$. This module $M$ is an assembly of stacked convolutional layers, complemented by a single fully connected layer, whose primary function is to align the feature dimensions across the two distinct modalities. For the training process, a GeForce RTX 4090 is employed, leveraging the Adam optimizer \cite{loshchilov2017decoupled}. The optimizer is initialized with a learning rate of 0.0001, and the training regimen extends for 10 epochs. 

\noindent\textbf{CERL. } In the training phase of the CERL model, the configuration was set to allow only $T_i$ to be learnable, while all other parameters remained fixed. This process utilized the GeForce RTX 4090 and employed a Stochastic Gradient Descent (SGD) optimizer \cite{robbins1951stochastic}. The initial learning rate for the optimizer was set to 0.1. Notably, the learning rate was decreased by a factor of 10 at the second, fourth, and sixth epochs, with the training extending over a total of 10 epochs.

\subsection{Comparison with baseline Methods}
\subsubsection{Quantitative Comparisons} 
We first present the results on the most widely used MEAD dataset in Table \ref{table: MEAD}. In the intra-ID setting, integrating the CDRL algorithm into both the ICface and NED baselines obtains evident improvement on all three metrics. Taking the NED baseline as an example, it reduces the average FAD from 2.108 to 1.026, the average LSE-D from 9.454 to 9.372, and increases the average CSIM from 0.831 to 0.914. These comparisons well suggest that CDRL can help to generate more real images with better emotion manipulation and audio-lip synchronization. Similar improvement in the three metrics can be observed when applying CDRL to the single-stage ICface baseline, well demonstrating its generalization abilities across different baseline methods. Cross-ID setting refers to more practical scenarios and integrating CDRL can also lead to performance improvements. It decreases the average FAD and LSE-D from 4.448 to 4.344 and from 9.906 to 9.351, with a relative decrement of 0.104 and 0.555, and increases the average CSIM from 0.773 to 0.792, with a relative increment of 0.019 when using NED baseline. 

\begin{figure*}[!t]
    \centering
    \begin{minipage}{0.87\textwidth}
        \centering
        \includegraphics[width=\textwidth]{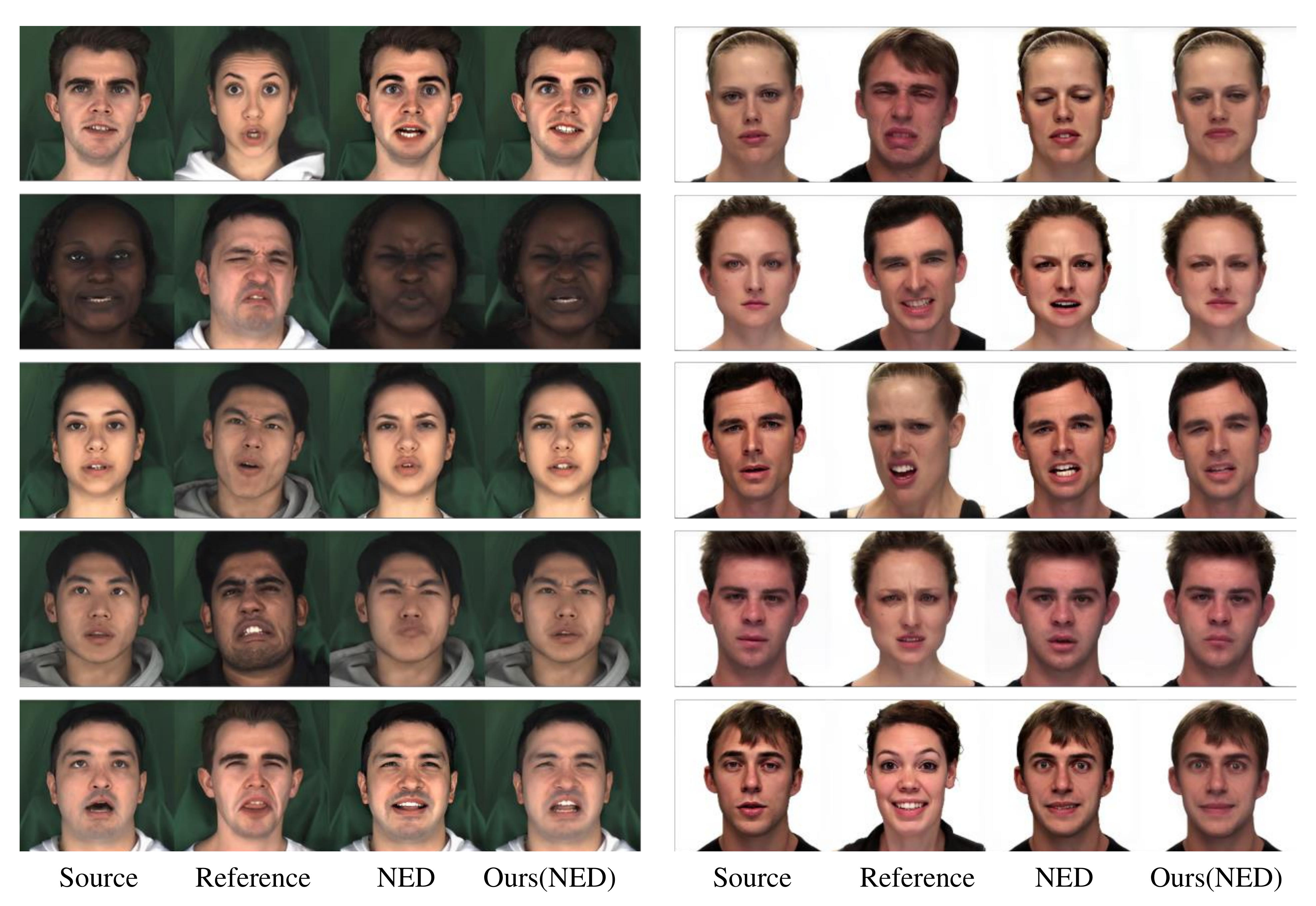}
        \vspace{-20pt}
        \caption{Qualitative comparisons of NED with and without the proposed algorithm. \textbf{Left half:} The samples are selected from the MEAD dataset. \textbf{Right half:} The samples are selected from the RAVDESS dataset.}
        \label{fig:MEAD_RAVDESS_NED}
    \end{minipage}
    \vspace{1em}
    \begin{minipage}{0.86\textwidth}
        \centering
        \includegraphics[width=\textwidth]{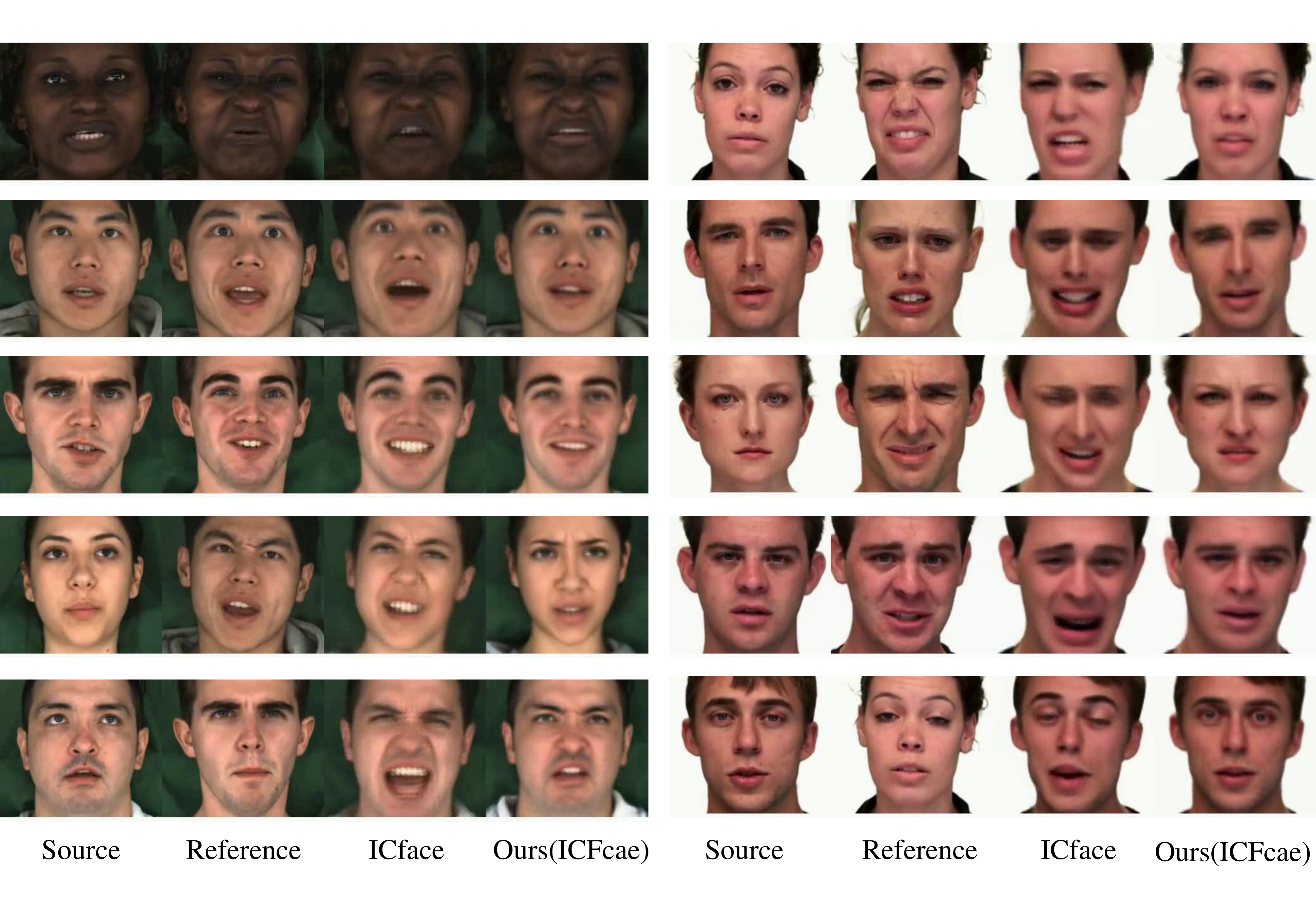}
        \vspace{-30pt}
        \caption{Qualitative comparisons of ICface with and without the proposed algorithm. \textbf{Left half:} The samples are selected from the MEAD dataset. \textbf{Right half:} The samples are selected from the MEAD dataset.}
        \label{fig:MEAD_RAVDESS_ICface}
    \end{minipage}
\end{figure*}

To demonstrate the generalization ability of the trained CDRL algorithm, we further present the performance comparisons on RAVDESS without retraining the CDRL algorithm. As shown in Table \ref{table: RAVDESS}, integrating the trained CDRL to both NED and ICface baselines also obtains evident improvement on all three metrics for the intra-ID and cross-ID settings. When using the NED baseline, it achieves average FAD and LSE-D decrements by 0.077 and 0.217, and CSIM increment by 0.012 for intra-ID settings. The improvement is even more evident for the cross-ID setting, decreasing average FAD and LSE-D from 5.412 to 4.816 and from 8.034 to 7.823, and increasing the CSIM from 0.760 to 0.771.



\begin{figure*}[htp]
  \centering
  \includegraphics[width=0.95\textwidth]{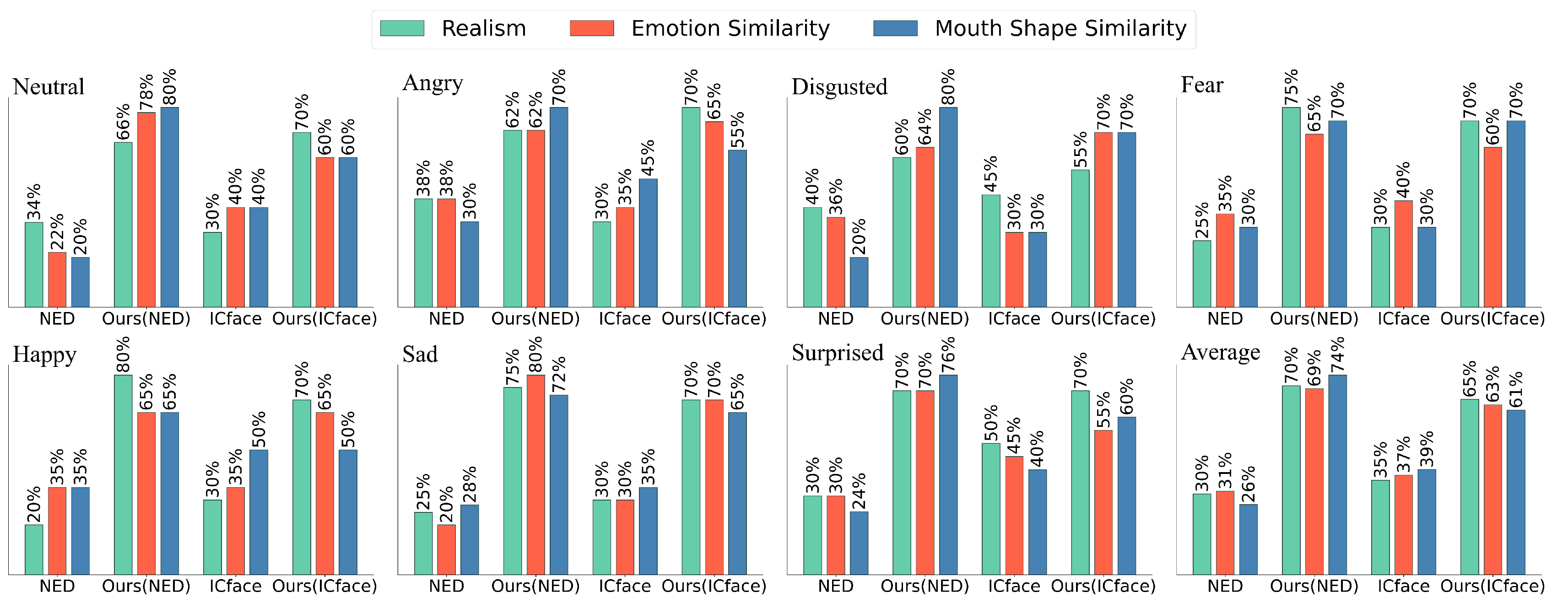} 
  \vspace{-10pt}
  \caption{\textcolor{black}{Realism, emotion similarity, and mouth shape similarity ratings of the user study on NED with and without CDRL, and on ICface with and without CDRL on the MEAD dataset.}}
  \label{fig:user-study-mead_comparison}
\end{figure*}

\begin{figure*}[htp]
  \centering
  \includegraphics[width=0.95\textwidth]{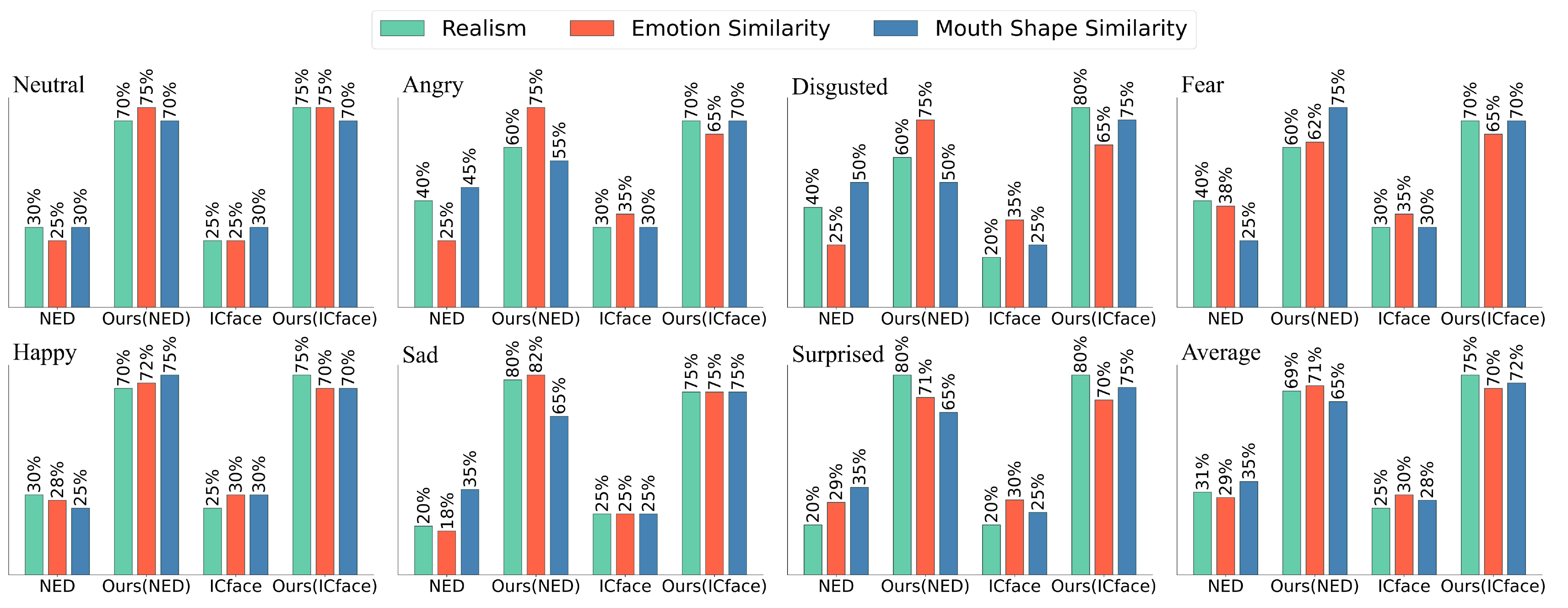} 
  \vspace{-10pt}
  \caption{\textcolor{black}{Realism, emotion similarity, and mouth shape similarity ratings of the user study on NED with and without CDRL, and on ICface with and without CDRL on the RAVDRSS dataset.}}
  \label{fig:user-study-ravdrss_comparison}
\end{figure*}

\subsubsection{Qualitative Comparisons.} 
In this section, we will present visual comparison results on the MEAD and RAVDESS datasets, showcasing NED with and without the proposed algorithm, and ICface with and without the proposed algorithm, as illustrated in Figures \ref{fig:MEAD_RAVDESS_NED} and \ref{fig:MEAD_RAVDESS_ICface}. Similar to the quantitative metrics, we will analyze the qualitative comparisons from three dimensions. 

\noindent\textbf{Realism. } NED employs a two-stage approach for emotion editing. The first stage predicts the edited 3DMM, and the second stage uses the 3DMM to generate the final result. Due to the lack of explicit supervision in the first stage of NED's training process, the predicted 3DMM cannot effectively maintain the original mouth shape, as shown in the third column of Figure \ref{fig:MEAD_RAVDESS_NED}. Additionally, the inaccuracies in 3DMM prediction can lead to distortions in the final rendering results, as illustrated in the third column of the fourth row of the left half of Figure \ref{fig:MEAD_RAVDESS_NED}. CDRL, by decoupling representations, provides explicit supervision for NED's training process, effectively maintaining the mouth shape during emotion editing and producing more realistic and natural results, as seen in the fourth column of Figure \ref{fig:MEAD_RAVDESS_NED}. ICface extracts emotion information from the reference's AU using a network for emotion editing. This method struggles to decouple emotion information from other information, resulting in some information distortion, as shown in the third column of Figure \ref{fig:MEAD_RAVDESS_ICface}. CDRL achieves more realistic editing effects by aligning the content representation between the source input and the generated output, as well as aligning the emotion representation between the reference input and the generated output, as seen in the fourth column of Figure \ref{fig:MEAD_RAVDESS_ICface}. 

\noindent\textbf{Emotion Similarity. } NED implicitly supervises the emotion editing at the 3DMM space, while ICface uses the reference as pseudo-labels for explicit supervision. However, neither method effectively decouples the emotion representation from other information, resulting in changes to the mouth shape during emotion editing, as shown in the third column of Figure \ref{fig:MEAD_RAVDESS_NED} and the third column of Figure \ref{fig:MEAD_RAVDESS_ICface}. Thanks to CERL, we can achieve emotion migration without altering the mouth shape, as illustrated in the fourth column of Figure \ref{fig:MEAD_RAVDESS_NED} and the fourth column of Figure \ref{fig:MEAD_RAVDESS_ICface}.  

\noindent\textbf{Lip-Audio Preserving Accuracy. } The training processes of NED and ICface lack explicit supervision for the mouth shape, leading to inconsistencies in the mouth shape before and after editing, as shown in the third column of Figure \ref{fig:MEAD_RAVDESS_NED} and the third column of Figure \ref{fig:MEAD_RAVDESS_ICface}. Thanks to CCRL, our results are more accurate in terms of mouth shape retention, as illustrated in the fourth column of Figure \ref{fig:MEAD_RAVDESS_NED} and Figure \ref{fig:MEAD_RAVDESS_ICface}. \emph{We will present some video comparisons for more direct comparison in} \href{https://jianmanlincjx.github.io/}{https://jianmanlincjx.github.io/}

\subsubsection{User Study}
In our web-based study, we assessed the performance of NED and ICface with and without the CDRL algorithm, focusing on three key metrics: realism, emotion similarity, and mouth shape similarity across seven basic emotions. We carefully selected 10 videos per emotion for both inter-identification and cross-identification, totaling 70 videos. Each of the 25 participants evaluated these aspects for every video. Our findings, detailed in Table \ref{fig:user-study-mead_comparison}, highlight that the CDRL algorithm significantly enhances the performance of both NED and ICface across the MEAD dataset. It consistently excels over the baseline across all emotions and metrics. On average, the integration of CDRL shows remarkable improvements: a 40\% increase in realism, 38\% in emotion similarity, and a notable 48\% in mouth shape similarity compared to the baseline NED on the MEAD dataset. Additionally, the integration of the CDRL algorithm with the ICface baseline markedly enhances realism, emotional congruence, and lip synchronicity. It achieves significantly higher ratings than the ICface baseline across all emotions. On average, the inclusion of the CDRL algorithm results in a 30\% higher rating in realism, 26\% more in emotion similarity, and 22\% more in mouth shape similarity compared to the NED baseline.

Similarly, Tables \ref{fig:user-study-ravdrss_comparison} present findings using the NED and ICface baselines on the RAVDESS dataset. Given that RAVDESS features a smaller pool of videos, we randomly selected 5 videos for each emotion, culminating in a total of 35 videos. This subset was then evaluated by the same cohort of 25 participants. Our findings indicate that the incorporation of the CDRL algorithm also yields notably higher ratings in all three aspects of the RAVDESS dataset.

\subsection{Ablation Study}
The above analyses and comparisons demonstrate the effectiveness of the proposed CDRL as a whole. Here, we conduct more experiments to analyze the actual contributions and provide more in-depth discussions for both the CCRL and CERL modules.  


\subsubsection{Analyses of CCRL}
\label{subsubsec:ccrl_analysis}
\begin{figure}[!t]
    \centering
    \includegraphics[width=0.48\textwidth]{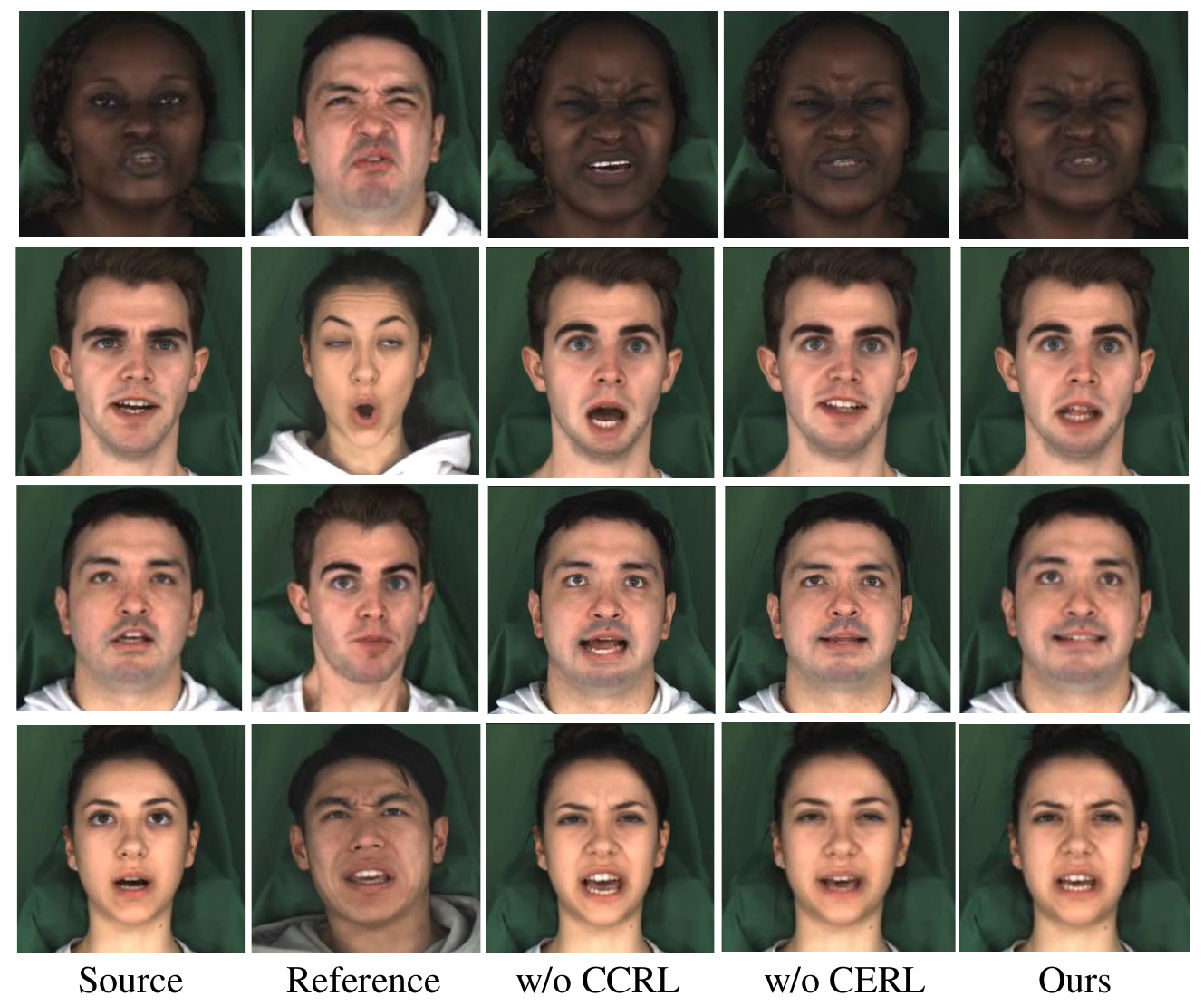}
    \caption{Qualitative comparisons of Ours, Ours w/o CCRL, and Ours w/o CERL using NED baseline.The samples are selected from the MEAD dataset.}
    \label{fig:ablation_MEAD_NED}
\end{figure}

\textcolor{black}{CCRL provides decoupled content representation to supervise the SPFEM model, and it is expected to help improve audio-lip synchronization. Here, we verify this point by comparing it with another baseline that removes the content representation supervision (namely ``Ours w/o CCRL'').} As shown in Table \ref{table: CCRL}, removing this supervision leads to a severe performance drop on LSE-D that measures audio-lip synchronization, an increment from 9.372 to 9.448 for intra-ID setting and from 9.351 to 9.883 for cross-ID setting. Besides, the CSIM metric that mainly reflects emotion similarity remains nearly unchanged. These results suggest the effectiveness of content representation in preserving mouth animation of spoken content. \textcolor{black}{Additionally, Fig . \ref{fig:ablation_MEAD_NED} presents visualization results comparing our method with and without CCRL. The removal of CCRL leads to noticeable discrepancies between the source and generated frame images. For example, in the first row, the mouth shape of the image generated by ``Ours w/o CCRL'' differs significantly from that of the source image. This underscores the importance of integrating CCRL into the training process of the SPFEM model.}

\begin{table}[h]
\centering
\small
\begin{tabular}{c|c|ccc}
\toprule
Settings & Methods & FAD$\downarrow$ & LSE-D$\downarrow$ & CSIM$\uparrow$ \\
\hline
\multirow{4}{*}{intra-ID} & NED & 2.108 & 9.454 & 0.831 \\
& Ours w/o CCRL  & 1.161 & 9.448 & 0.911 \\
& CCRL w/o audio  & 1.214 & 9.446 & 0.909 \\
& Ours & 1.026 & 9.372 & 0.914 \\
\hline
\multirow{4}{*}{cross-ID} & NED & 4.448 & 9.906 & 0.773 \\
& Ours w/o CCRL & 4.401 & 9.883 & 0.786 \\
& CCRL w/o audio  &4.411 & 9.812 & 0.785 \\
& Ours & 4.344 & 9.351 & 0.792 \\
\bottomrule
\end{tabular}
\caption{FAD, LSE-D, and CSIM of Ours, Ours CCRL w/o audio, Ours w/o CCRL, and NED baseline.}
\label{table: CCRL}
\end{table}

\begin{table}[h]
\centering
\small
\begin{tabular}{c|c|ccc}
\toprule
Settings & Methods & FAD$\downarrow$ & LSE-D$\downarrow$ & CSIM$\uparrow$ \\
\hline
\multirow{4}{*}{inter-ID} & NED & 2.108 & 9.454 & 0.831 \\
& CCRL w/o emotion  & 1.287 & 9.399 & 0.901 \\
& Ours & 1.026 & 9.372 & 0.914 \\
\hline
\multirow{4}{*}{cross-ID} & NED & 4.448 & 9.906 & 0.773 \\
& CCRL w/o emotion & 4.393 & 9.382 & 0.791 \\
& Ours & 4.344 & 9.351 & 0.792 \\
\bottomrule
\end{tabular}
\caption{FAD, LSE-D, and CSIM of Ours, Ours CCRL w/o emotion and NED baseline.}
\label{table: CCRL_}
\end{table}

CCRL exploits audio as content prior to guiding learning content information. Here, we further conduct an experiment (namely ``CCRL w/o audio'') that excludes the audio and simply uses the images to learn content representation via identical contrastive learning. The comparison results are presented in Table \ref{table: CCRL}. In the cross-ID setting, it increases the LSE-D from 9.351 to 9.812, an evident performance degradation on audio-lip synchronization.

During the training process of CCRL, we incorporate emotion-aware contrastive learning, which entails careful consideration of the emotional element in constructing positive and negative samples. Specifically, the positive samples comprise two images with identical spoken content but differing emotions. In contrast, the negative samples consist of two images sharing the same emotion but with different spoken content. This emotion-aware contrastive learning is designed to decouple emotion-independent content information from images more effectively. To validate the effectiveness of this, we devise an experiment (``CCRL w/o emotion''), in which the negative samples are constructed solely as images with differing spoken content. The comparison results are presented table \ref{table: CCRL_}.

From the table \ref{table: CCRL_}, it is evident that even without considering the element of emotion in constructing negative samples, ``CCRL w/o emotion'' still plays a guiding role in NED. Compared to NED itself, ``CCRL w/o emotion'' shows a significant improvement in the inter-ID setting, with FAD and LSE-D decreasing from 2.108 to 1.287 and from 9.454 to 9.399, respectively, and CSIM increasing from 0.831 to 0.901. Similarly, there is a noticeable enhancement in the Cross-ID setting. However, compared to ``Ours'', ``CCRL w/o emotion'' exhibits a substantial increase in LSE-D, rising from 9.351 to 9.382. This indicates that thoroughly considering the element of emotion in the construction of negative samples can further decouple emotion-independent content information from the image, thereby promoting lip synchronization. This also underscores the effectiveness of emotion-aware contrastive learning in maintaining high-quality lip sync.

\begin{figure}[!t]
    \centering
    \includegraphics[width=0.48\textwidth]{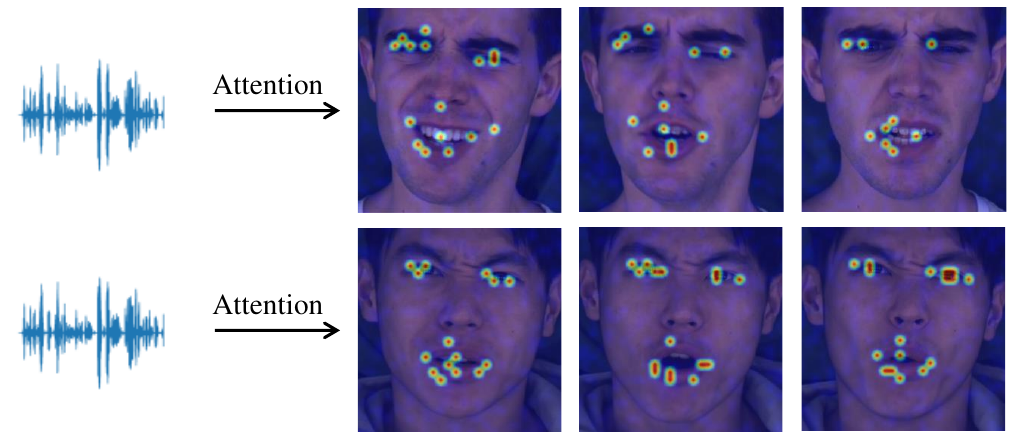}
    \caption{Visualization of attention maps in the form of heatmaps.}
    \label{fig:attention_image}
\end{figure}

CCRL introduces audio to guide learning content-related representation. Here, we further map the attention weight derived from the matrix multiplication of query and key back to the original images. As shown in Fig. \ref{fig:attention_image}, we find the activation regions mainly located in the mouth and eye areas. In human communication, both the eyes and the mouth are primary areas for conveying content. Although audio itself is not strongly associated with the eyes, through appropriate training strategies, we can guide CCRL to use audio to focus on areas in the image closely related to the content, including the mouth and eyes. These results further demonstrate the benefit of decoupled content representation learning. 

\begin{table}[!t]
\centering
\small
\begin{tabular}{c|c|ccc}
\toprule
Settings & Methods & FAD$\downarrow$ & LSE-D$\downarrow$ & CSIM$\uparrow$ \\
\hline
\multirow{4}{*}{intra-ID} & NED & 2.108 & 9.454 & 0.831 \\
& Ours w/o CERL  & 1.124 & 9.384 & 0.837 \\
& Ours CERL CA  & 1.287 & 9.389 & 0.899 \\
& Ours & 1.026 & 9.372 & 0.914 \\
\hline
\multirow{4}{*}{cross-ID} & NED & 4.448 & 9.906 & 0.773 \\
& Ours w/o CERL & 4.384 & 9.393 & 0.774 \\
& Ours CERL CA  & 4.395 & 9.388 & 0.789 \\
& Ours & 4.344 & 9.351 & 0.792 \\
\bottomrule
\end{tabular}
\caption{FAD, LSE-D, and CSIM of Ours, Ours CERL CA, Ours w/o CERL, and NED baseline.}
\label{table: CERL}
\end{table}

\begin{table}[!t]
\centering
\small
\begin{tabular}{c|c|ccc}
\toprule
Settings & Methods & FAD$\downarrow$ & LSE-D$\downarrow$ & CSIM$\uparrow$ \\
\hline
\multirow{4}{*}{inter-ID} & NED & 2.108 & 9.454 & 0.831 \\
& Ours CERL w/o AG & 1.147 & 9.387 & 0.888 \\
& Ours & 1.026 & 9.372 & 0.914 \\
\hline
\multirow{4}{*}{cross-ID} & NED & 4.448 & 9.906 & 0.773 \\
& Ours CERL w/o AG & 4.388 & 9.393 & 0.779 \\
& Ours & 4.344 & 9.351 & 0.792 \\
\bottomrule
\end{tabular}
\caption{FAD, LSE-D, and CSIM of Ours, Ours CERL w/o AG and NED baseline.}
\label{table: CERL_}
\end{table}

\subsubsection{Analyses of CERL}
\label{subsubsec:cerl_analysis}
\textcolor{black}{CERL learns emotion representation to help better modify the emotional states. To verify its contribution, we also carry out an experiment of removing it for comparisons (namely ``Ours w/o CERL'').} As exhibited in Table \ref{table: CERL}, CSIM drops from 0.914 to 0.837 for the intra-ID setting and from 0.792 to 0.774 for the cross-ID setting, a severe degradation in emotion alignment. \textcolor{black}{Similarly, we also present some visualization results of Ours with and without CERL. Obviously, integrating CERL can obtain better expression manipulation. As shown in the second row of Fig. \ref{fig:ablation_MEAD_NED}, the reference expression is ``surprised''. The proposed algorithm successfully modifies the expression to ``surprised'' whereas removing CERL results in a less effective modification. Both qualitative and quantitative comparisons highlight the significant impact of CERL on expression manipulation.}

CERL exploits a simple dot product between the corresponding emotion prior and the feature vector of a given image, while CCRL introduces a cross-attention mechanism. To analyze this point, we carry out an experiment using the identical cross-attention mechanism to fuse emotion prior and image features (namely ``Ours CERL CA''). As shown in Table \ref{table: CERL}, it performs slightly worse than that using a simple dot product. One possible reason may be visual-language model inherently uses dot product to compute visual-language alignment, and incurring other mechanisms may destroy the alignment. Notably, ``Ours CERL CA'' still shows clear CSIM improvement over the NED baseline, suggesting that emotion representation is effective across different mechanisms.

CERL operates by distilling emotion prior from images for each text embedding set $T_i$. During the training phase of CERL, we introduce emotion-augmented contrastive learning. This involves utilizing CLIP to filter out the most expressively emotional images from the MEAD dataset to serve as training samples. In this context, we explore an experiment of not utilizing emotion-augmented contrastive learning (termed as ``Ours CERL w/o AG'') and instead, rely solely on randomly obtaining samples from the MEAD dataset for training purposes. As demonstrated in the Table \ref{table: CERL_}, compared to NED itself, the non-utilization of emotion-augmented contrastive learning(``Ours CERL w/o AG'') leads to improvements across all metrics. In the cross-id setting, FAD and LSE-D experienced relative changes of 0.06 and 0.513, respectively, and CSIM increased from 0.773 to 0.779. Notable enhancements are also observed in the inter-ID setting. However, compared to ``Ours'', ``Ours CERL w/o AG'' exhibits an increase of 0.042 in LSE-D and a decrease of 0.013 in CSIM. A potential reason for this might be that the training set for ``Ours'' comprises images with strong emotional expressiveness. \textcolor{black}{This could enable the network to learn the underlying differences between emotions and better decouple emotional representations that are independent of content, thereby enhancing emotional expressiveness while improving lip synchronization.}

\begin{figure}[!t]
    \centering
    \includegraphics[width=0.3\textwidth]{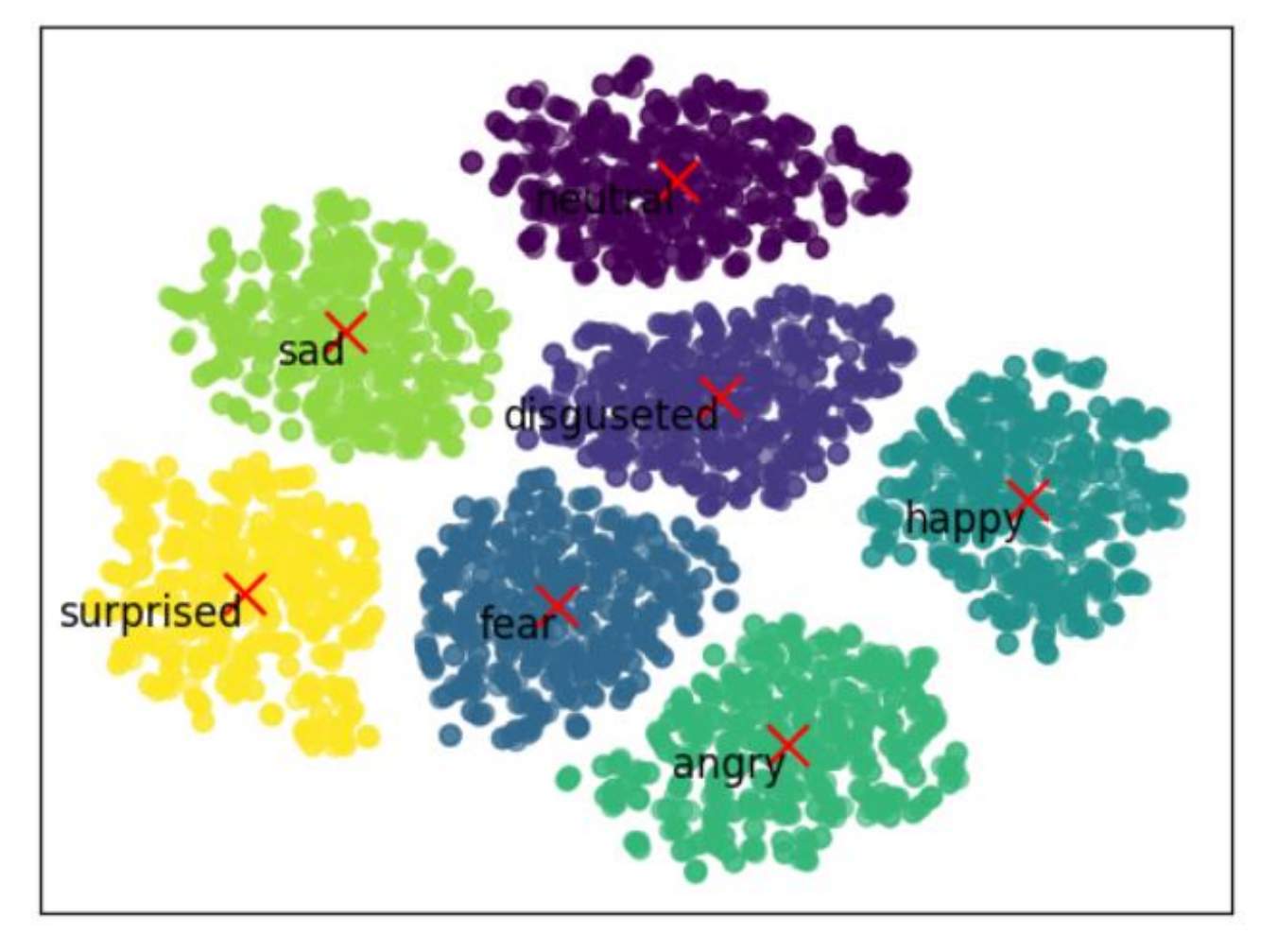}
    \caption{t-SNE visualization of emotion representation for the ``neutral", ``happy", ``angry", ``surprised", ``fear", ``sad", and ``disgusted".}
    \label{fig:emotion-representation}
\end{figure}

To delve deeper into CERL, we visualize the emotion representations as shown in Fig. \ref{fig:emotion-representation}. The emotion representations are obtained through the direct dot product between emotion priors and corresponding emotional image features. We find that representations of the same emotion gather together, while those of different emotions distance themselves from each other. This demonstrates that CERL can effectively learn emotion-related representations, providing strong supervisory signals.

We also conducted an ablation experiment utilizing fixed prompts such as ``happy'', and a CLIP text encoder to extract their features as an emotional prior, without prompt tuning. But it yielded unsatisfactory results. One possible reason is that simple prompts such as ``happy'', really contain emotional semantic information. However, due to CLIP’s vast image-text training data, such a prompt is not specific to the speaker’s image and thus cannot provide effective supervision. So we get the emotion prior through prompt tuning, which is more reasonable and effective.

\subsection{\textcolor{black}{Integration with Other Supervision Signals}}
\textcolor{black}{ASCCL \cite{chen2024learning} is a newly proposed method, offering supervision for the SPFEM model, but from a distinctly different perspective compared to CDRL. Specifically, ASCCL \cite{chen2024learning} explores spatial correlations from paired data and uses these correlations as additional supervision. In contrast, CDRL directly learns content and emotional information from the source and reference videos, respectively. This mechanism better fits the SPFEM task as it requires modifying the emotion according to the reference video and meanwhile maintaining the mouth movement of the source video.}

\textcolor{black}{Furthermore, these two methods provide additional supervision from distinct perspectives and appear to be complementary. To empirically validate their complementary nature, we integrated both ASCCL \cite{chen2024learning} and CDRL into the current SPFEM models, resulting in further performance improvements, as shown in Table \ref{table: Integration with ASCCL}. Notably, incorporating ASCCL into CDRL led to a decrease of 0.092 and 0.122 in the FAD and LSE-D metrics, respectively, while showing a 0.002 increase in the CSIM metric under the cross-ID settings. A similar pattern is observed in the inter-ID setting. Additionally, the combination of ASCCL and CDRL outperforms ASCCL alone by 0.223, 0.016, and 0.019 in the inter-ID setting, and by 0.012, 0.009, and 0.003 in the cross-ID setting for the FAD, LSE-D, and CSIM metrics, respectively. This demonstrates that CDRL and ASCCL \cite{chen2024learning} are mutually reinforcing and can be combined to enhance the SPFEM task.}

\begin{table}[!t]
\centering
\small
\begin{tabular}{c|c|ccc}
\toprule
Settings & Methods & FAD$\downarrow$ & LSE-D$\downarrow$ & CSIM$\uparrow$ \\
\hline
\multirow{4}{*}{inter-ID} 
& ASCCL & 1.234  & 9.340 & 0.900 \\
& CDRL & 1.026 & 9.372 & 0.914 \\
& CDRL w/ ASCCL & 1.011 & 9.324 & 0.919 \\
\hline
\multirow{4}{*}{cross-ID} 
& ASCCL & 4.264 & 9.238 & 0.791 \\
& CDRL & 4.344 & 9.351 & 0.792 \\
& CDRL w/ ASCCL & 4.252 & 9.229 & 0.794 \\
\bottomrule
\end{tabular}
\caption{\textcolor{black}{Performance comparison of ASCCL \cite{chen2024learning}, CDRL, and their combination (CDRL w/ ASCCL) on the NED \cite{papantoniou2022neural} baseline using the MEAD dataset \cite{wang2020mead}. Integrating ASCCL and CDRL into the NED training process demonstrates further improvements in FAD, LSE-D, and CSIM metrics.}}
\label{table: Integration with ASCCL}
\end{table}

\subsection{\textcolor{black}{Limitation}}
\textcolor{black}{CDRL is pre-trained on the MEAD dataset, and we conducted experiments on the RAVDESS dataset without retraining CDRL. The experiments demonstrate that CDRL possesses certain pre-adaptation capabilities. However, in some cases, it still fails to achieve ideal results, such as in the right half of the last example in Fig. \ref{fig:MEAD_RAVDESS_NED}, where the results are not ideal, particularly in accurately transferring details like teeth. In future work, we plan to further enhance the pre-adaptation capabilities of CDRL, for example, by using adversarial training to decouple domain-independent representations more effectively, thereby improving CDRL’s generalization ability.}

\section{Conclusion}
This work presents a Contrastive Decoupled Representation Learning (CDRL) algorithm, which learns decoupled content and emotion representation as more direct and accurate supervision signals to facilitate Speech-preserving Facial Expression Manipulation (SPFEM). It consists of Contrastive Content Representation Learning (CCRL) and Contrastive Emotion Representation Learning (CERL) modules, in which the former exploits audio as content prior to learning emotion-independent content representation while the latter introduces large-scale visual-language model to learn emotion prior, which is then used to guide learning content-independent emotion representation. During CCRL and CERL learning, we use contrastive learning as the objective loss to ensure that content and emotion representation merely contain content and emotion information, respectively. During SPFEM model training, the decoupled content and emotion representation are used in the generation process, ensuring more accurate emotional manipulation together with audio-lip synchronization. Extensive experiments and evaluations across various benchmarks have demonstrated the effectiveness of the proposed CDRL algorithm.

\section*{Acknowledgment}
This work was supported in part by National Natural Science Foundation of China (62206060, 61972163), Natural Science Foundation of Guangdong Province \\ (2023A1515012561, 2022A1515011555, SL2022A04J01626, 2023A1515012568), Guangdong Provincial Key Laboratory of Human Digital Twin (2022B1212010004).

\noindent{\textbf{Code availability.} } \quad All trained models and codes are publicly available on GitHub: \url{https://github.com/jianmanlincjx/CDRL}.

\noindent{\textbf{Data availability.} } \quad The data that support the finding of this study are openly available at the following URL: \url{https://github.com/uniBruce/Mead}, \url{https://paperswithcode.com/dataset/ravdess}.


\bibliographystyle{spmpsci}      
\bibliography{referencce} 

\end{document}